\documentclass[a4paper,10pt,twocolumn]{article}

\usepackage[english]{babel}
\usepackage[utf8]{inputenc}
\usepackage[T1]{fontenc}

\newcommand{\norm}[2]{\left\Vert  {#1} \right\Vert _{#2}}

\usepackage[top=1.5cm, left=1.5cm, right=1.5cm, bottom=1.5cm]{geometry}

\renewenvironment{abstract}{\bf\small {\em\ Abstract---}}{}

\usepackage{amsfonts,amssymb,amsmath,amsthm}
\usepackage{subfigure}
\usepackage{graphicx}
\usepackage[footnotesize]{caption}

\usepackage{amsfonts,amssymb,amsmath,amsthm}

\title{Heuristics for Efficient Sparse Blind Source Separation}

\author{Christophe Kervazo$^1$, J\'{e}r\^{o}me Bobin$^1$ and C\'{e}cile Chenot$^2$.\\
  \footnotesize $^1$CEA Saclay, Gif-sur-Yvette, 91191 cedex, France \ $^2$Institute for Digital Communications, University of Edinburgh, UK\
  } \date{\empty} 

\begin{document}

\maketitle

\begin{abstract} 
Sparse Blind Source Separation (sparse BSS) is a key method to analyze multichannel data in fields ranging from medical imaging to astrophysics. However, since it relies on seeking the solution of a {\it non-convex} penalized matrix factorization problem, its performances largely depend on the optimization strategy.
In this context, Proximal Alternating Linearized Minimization (PALM) has become a standard algorithm which, despite its theoretical grounding, generally provides poor practical separation results. In this work, we propose a novel strategy that combines a heuristic approach with PALM. We show its relevance on realistic astrophysical data.
\end{abstract}

\section{Introduction}
\subsection{Blind Source Separation problem}
In the BSS \cite{comon2010handbook} framework, the data are composed of $m$ observations, each of which has $t$ samples. These observations are supposed to be some linear combinations of $n$ sources. The objective of BSS is to retrieve the sources as well as the mixing coefficients. In matrix form, the goal is therefore to find two matrices $\mathbf{S}$ (of size $n\times t$) and $\mathbf{A}$ (of size $m \times n$), called respectively the source and the mixing matrices, such that: $\mathbf{X = AS + N}$, where $\mathbf{X}$ (of size $m\times t$) is the observation matrix that is corrupted with some unkwown noise $\mathbf{N}$. Since it requires tackling an ill-posed unsupervised matrix factorization problem, further assumptions are needed, including the statistical independance of the sources (ICA - \cite{comon2010handbook}), the non-negativity of $\mathbf{A}$ and $\mathbf{S}$ \cite{gillis2012accelerated}. In this work, we will focus on the sparsity of the sources  \cite{bobin20085,zibulevsky2001blind,bronstein2005sparse,li2006underdetermined}. In this framework, sparse BSS will aim at finding a (local) minimum of:
\begin{equation}
\small
\min_{\mathbf{A,S}} \frac{1}{2} \norm{\mathbf{X-AS}}{F}^{2} 
+ \iota_{\{\forall i; \|{\bf A}^i \|_2^2 \leq 1\}}(\bf A)
+ \norm{R_\mathbf{S}\odot(\mathbf{S\Phi_S^T})}{1}
\label{funcGMCA}
\end{equation}
The first data fidelity term promotes a faithful reconstruction of the data. The use of the Froebenius norm $\norm{.}{F}$ stems from the assumption of a Gaussian noise $\mathbf{N}$. The second term ensures that the columns of $\mathbf{A}$ are all in the $\ell_2$ ball. This avoids degenerated $\mathbf{A}$ and $\mathbf{S}$ ($\norm{\mathbf{A}}{ } \rightarrow \infty$ and $\norm{\mathbf{S}}{ } \rightarrow 0$). The last term involving the Hadamard product $\odot$ enforces a $\ell_1$ sparsity constraint in a transformed domain $\mathbf{\Phi_S}$ (here, $\mathbf{\Phi_S}$ will be the starlet transform \cite{starck2010sparse}). $\mathbf{R_\mathbf{S}}$ controls the trade-off between the data fidelity and the sparsity terms. It can be decomposed into $\mathbf{R_\mathbf{S}} = \mathbf{\mathbf{\Lambda_S}} \mathbf{W}$ where $\mathbf{\mathbf{\Lambda_S}}$ is a diagonal matrix of the regularization parameters $\lambda_1, \lambda_2,...,\lambda_n$ and $\mathbf{W}$ is a matrix used to introduce individual penalization coefficients in the context of reweighted $\ell_1$ \cite{candes2008enhancing}.

\subsection{Sparse BSS in practice}
\label{thrdStrat}
Since sparse BSS requires solving a penalized matrix factorization problem, the separation quality strongly depends on the optimization strategy. Different algorithmic frameworks have been used so far: projected Alternate Least-Square (ALS - \cite{gillis2012accelerated}), PALM \cite{bolte2014proximal} and Block-Coordinate Descent (BCD - \cite{tseng2001convergence}, which is not studied here due to a high computational burden):
\begin{itemize}
	\item \emph{PALM algorithm}: PALM is an iterative algorithm, which alternates at each iteration between a proximal gradient step on $\mathbf{A}$ and $\mathbf{S}$.  PALM is proved to converge to a local minimum of (\ref{funcGMCA}) under mild conditions \cite{bolte2014proximal}.
	\item \emph{GMCA algorithm}: the Generalized Morphological Component Analysis (GMCA - \cite{bobin20085}) algorithm is based on projected ALS. At each iteration (k), the gradient step appearing in PALM is replaced by a multiplication by a pseudo-inverse. Compared to PALM, GMCA cannot be proved to converge. Furthermore, even when stabilizing the output of GMCA is not really minimizing (\ref{funcGMCA}). 
\end{itemize}

In practice, the solution of sparse BSS methods is highly sensitive to the initial point and the values of the regularization parameters, which are generally tricky to tune without any first guess of the solution.
Thanks to heuristics, GMCA benefits from an automatic thresholding strategy (see Sec.\ref{sec:DiffParamPALMGMCA}) which has been empirically shown to improve its robustness with respect to the initialization and give good estimations of $\mathbf{A^*}$ and $\mathbf{S^*}$. Such heuristics do not exist in PALM.

\subsection{Contributions}
While PALM is theoretically well rooted and yields rather fast minimization schemes (in contrast to BCD), it generally provides poor separation results. We show how PALM-based implementations can benefit from the information provided by heuristic approaches which are in practice more robust. The robustness of the proposed combined strategy is demonstrated on realistic astrophysical data.

\section{Complexity of introducing heuristics in PALM}
\label{HeurInPalm}
Building on the automatic thresholding strategy of GMCA, the goal of this part is to try to derive one for PALM.
\subsection{Automatic parameter choice in GMCA}\label{sec:DiffParamPALMGMCA}
In GMCA, the threshold choice is performed computing the Median Absolute Deviation (MAD) for each currently estimated source $\mathbf{\hat{S}}$. The corresponding threshold (which changes during the iterations) is set to a multiple $k$ of this value:
\begin{equation}
\small
\left( \lambda_1,\lambda_2,...,\lambda_n \right)^T
= k\times \mathtt{MAD}(\mathbf{\hat{S}})
\label{eq:kmad}
\end{equation}

In this equation and in the remaining of this section, ${\bf \Phi_S^T}$ was supposed to be the identity matrix without loss of generality. $k$ is a positive number and the $\mathtt{MAD}$ operator is computed row-by-row with $\mathtt{MAD}(z) = \mathtt{median}(|z - \mathtt{median}(z)|)$ for $z\in \mathbb{R}^t$. 
Using this strategy, the threshold choice can be interpreted as a dense noise removal.
\subsection{Introducing heuristics in PALM}
\label{MotivHeur}
Let us assume that the thresholds in PALM are computed the same way as in GMCA through Eq.~(\ref{eq:kmad}) and that during the iterations the algorithm finds estimates that are close to \emph{both} the true matrices $\mathbf{A}^*$ and $\mathbf{S}^*$. Then, due to the assumption of $\mathbf{S}^*$ sparsity:
\begin{equation}
\small
\left( \lambda_1,\lambda_2,...,\lambda_n\right)^T
\simeq k\times \frac{\gamma}{\norm{\mathbf{A}^{*T} \mathbf{A}^{*}}{2}}\mathtt{MAD}(\mathbf{A}^{*T} \mathbf{N})
\end{equation}
Therefore, using the MAD enables a thresholding of a projection of the noise $\mathbf{N}$, which yields a similar interpretation as in GMCA. \emph{However, this interpretation requires that the algorithm finds good estimates of $\mathbf{A^*}$ and $\mathbf{S^*}$ during the iterations} (contrary to GMCA, which does not require to find a good estimate of $\mathbf{S^*}$ for the interpretation to be verified). In practice, using directly this strategy within a PALM algorithm therefore does not yield good estimates because when initialized with a random initialization the algorithm never becomes close to such good estimates.

\section{Combining GMCA and PALM: a hybrid strategy}
In this part, we propose to combine the best of GMCA and PALM in a two step approach. The algorithm comprehends a warm-up stage, in which GMCA is performed, followed by a refinement stage during which PALM is performed retaining as much information as possible coming from the warm-up stage.

\subsection{Motivation and full description of the algorithm}
Our approach is motivated by several remarks: i) \emph{PALM theoretical background:} in particular, the 2-step algorithm is thus proved to converge and, once the weights $\mathbf{W}$ estimated by the warm-up stage, it truly looks for a critical point of Eq.~(\ref{funcGMCA}); ii) \emph{GMCA robustness with regards to initialization}, which will lead to a robust 2-step algorithm; iii) \emph{Benefit from GMCA solution:}
	\begin{itemize}
		\item Since \emph{both} $\mathbf{A_{GMCA}}$ and $\mathbf{S_{GMCA}}$ are close to $\mathbf{A}^*$ and $\mathbf{S}^*$, they can be used to derive the thresholds using the MAD according to the previous section.
		\item $\mathbf{S_{GMCA}}$ should already give a good approximation of the most prominent peaks. This can be exploited in the refinement stage through the introduction of reweighted L1 \cite{candes2008enhancing} using the reweighting matrix $\mathbf{W}$ in problem (\ref{funcGMCA}): $\mathbf{W}_i^j = \frac{\epsilon}{\epsilon + \frac{|\mathbf{\hat{S}}^j_i|}{\norm{\mathbf{\hat{S}}_i}{\infty}}}$,
		with $\epsilon$ a small constant, $\mathbf{W}_i^j$ the coefficient of $\mathbf{W}$ corresponding to the {\it i}\textsuperscript{th} line and {\it j}\textsuperscript{th} column and $\mathbf{\hat{S}}_i$ the {\it i}\textsuperscript{th} line of $\mathbf{\hat{S}}$.
	\end{itemize}

These remarks lead to the following 2-step algorithm:

{\hspace{-0.8cm}
	\fbox{
		\parbox{\linewidth}{
			Input : $\mathbf{X}$ (data matrix)
			\begin{itemize}
				\item Random initialization $\mathbf{A_0}$ and $\mathbf{S_0}$
				\item \emph{Warm-up stage}: \\$\mathbf{A_{GMCA}}$, $\mathbf{S_{GMCA}}$ = GMCA($\mathbf{X}$,$\mathbf{A_0}$,$\mathbf{S_0}$)
				\item \emph{Refinement stage}: \\$\mathbf{A_{PALM}}$, $\mathbf{S_{PALM}}$ = PALM($\mathbf{X}$,$\mathbf{A_{GMCA}}$,$\mathbf{S_{GMCA}}$)\\ The initialization, thresholding strategy and reweighting information come from the warm-up stage.
			\end{itemize}
		}
	}
}
%

\section{Experiment on realistic data}

\subsection{Data description and experimental protocol}
The goal of this part is to apply our algorithm on realistic data to show its efficiency.
The $n = 2$ sources come from simulations obtained from real data of Cassiopeia A supernova remnant.
They each consists in a 2D image of resolution $t = $ 128 $\times$ 128 pixels, supposed to be approximately sparse in the starlet domain. The mixing is performed through a $\mathbf{A}$ matrix drawn randomly following a standard normal distribution and modified to have unit columns. Its condition number is $C_d = 10$. There is $m = n$ observations. To increase the realism of the data and further test the algorithm, we tried three relatively low SNR values: 10, 15 and 20 dB. $k$ is set to 3, which corresponds to a classical hypothesis in terms of Gaussian noise removal.
\subsection{Empirical results}
\begin{table}[!t]
	\caption{Average $C_A$ (10 different initializations) for 3 SNR values and 5 algorithms.}
	\label{fig_kmadHybChandra}
	\centering
	
	\begin{tabular}{llll}
		
		& 10 dB & 15dB & 20 dB\\
		
		2 step & {\bf 11.57} & {\bf 16.92} & {\bf 22.09}\\
		PALM &  9.38 & 10.94 & 11.01\\
		GMCA &  8.87 & 12.15 & 15.87\\
		EFICA & -6.92 & 5.11 & 9.41\\
		RNA   &  -6.68 & -5.49 & 6.27\\
		
	\end{tabular}
\end{table}
The mean of of the mixing matrix criterion $C_A$ \cite{bobin2015sparsity} is displayed in Table~\ref{fig_kmadHybChandra}. The 2-step approach always achieve better results than the two classical BSS algorithms with which we performed the comparison, namely Relative Newton Algorithm (RNA) and Eficient FastICA (EFICA). It also outerpeforms both GMCA and a PALM using directly the MAD heuristic, being always better by at least 2 dB than the best of them.\\
In addition to the results displayed in Fig.~\ref{fig_kmadHybChandra}, the standard deviation of $C_A$ over different initializations is almost 0, which shows the robustness of the algorithm.

\section*{Conclusion}
In this work, we introduce a 2-step strategy combining PALM with robust heuristic methods such as GMCA. Beyond improving the robustness of PALM-based implementations with respect to initialization, the regularization parameters can be automatically set in the proposed approach. Numerical experiments on realistic data demonstrate a high separation quality and good robustness on mixings with low SNR.

\section*{Acknowledgment}
This work is supported by the European Community through the grant LENA (ERC StG - contract no. 678282).


\end{document}